\documentclass[10pt,twocolumn,letterpaper]{article}

\usepackage[pagenumbers]{wacv} % To force page numbers, e.g. for an arXiv version

\definecolor{wacvblue}{rgb}{0.21,0.49,0.74}
\usepackage[pagebackref,breaklinks,colorlinks,allcolors=wacvblue]{hyperref}

\usepackage{graphics} % for pdf, bitmapped graphics files
\usepackage{epsfig} % for postscript graphics files
\usepackage{mathptmx} % assumes new font selection scheme installed
\usepackage{times} % assumes new font selection scheme installed
\usepackage{amsmath} % assumes amsmath package installed
\usepackage{amssymb}  % assumes amsmath package 
\usepackage{threeparttable}
\usepackage{amssymb}
\usepackage{newfloat}
\usepackage{listings}
\usepackage{float}

\def\wacvPaperID{} % *** Enter the WACV Paper ID here
\def\confName{WACV}
\def\confYear{2027}

\title{PDB: Not All Drivers Are the Same – A Personalized Dataset for Understanding Driving Behavior}

\author{
Chuheng Wei$^{1,2}$ 
\qquad   
Ziye Qin$^1$\qquad 
Siyan Li$^1$\qquad  
Ziyan Zhang$^1$\qquad
Xuanpeng Zhao$^1$\qquad \\
Amr Abdelraouf$^3$ \qquad
Rohit Gupta$^3$ \qquad 
Kyungtae Han$^3$ \qquad
Ziran Wang$^2$ \qquad \\
Matthew J. Barth$^1$ \qquad
Guoyuan Wu$^1$\qquad \\
$^1$ University of California, Riverside\qquad $^2$ Purdue University\\
$^3$ InfoTech Labs, Toyota Motor North America\\
{\tt\small chuheng.wei@email.ucr.edu}
}

\begin{document}
\makeatletter
\g@addto@macro\@maketitle{
  \begin{figure}[H]
  \vspace{-1em}
  \setlength{\linewidth}{\textwidth}
  \setlength{\hsize}{\textwidth}
  \centering
    \includegraphics[width=1\linewidth]{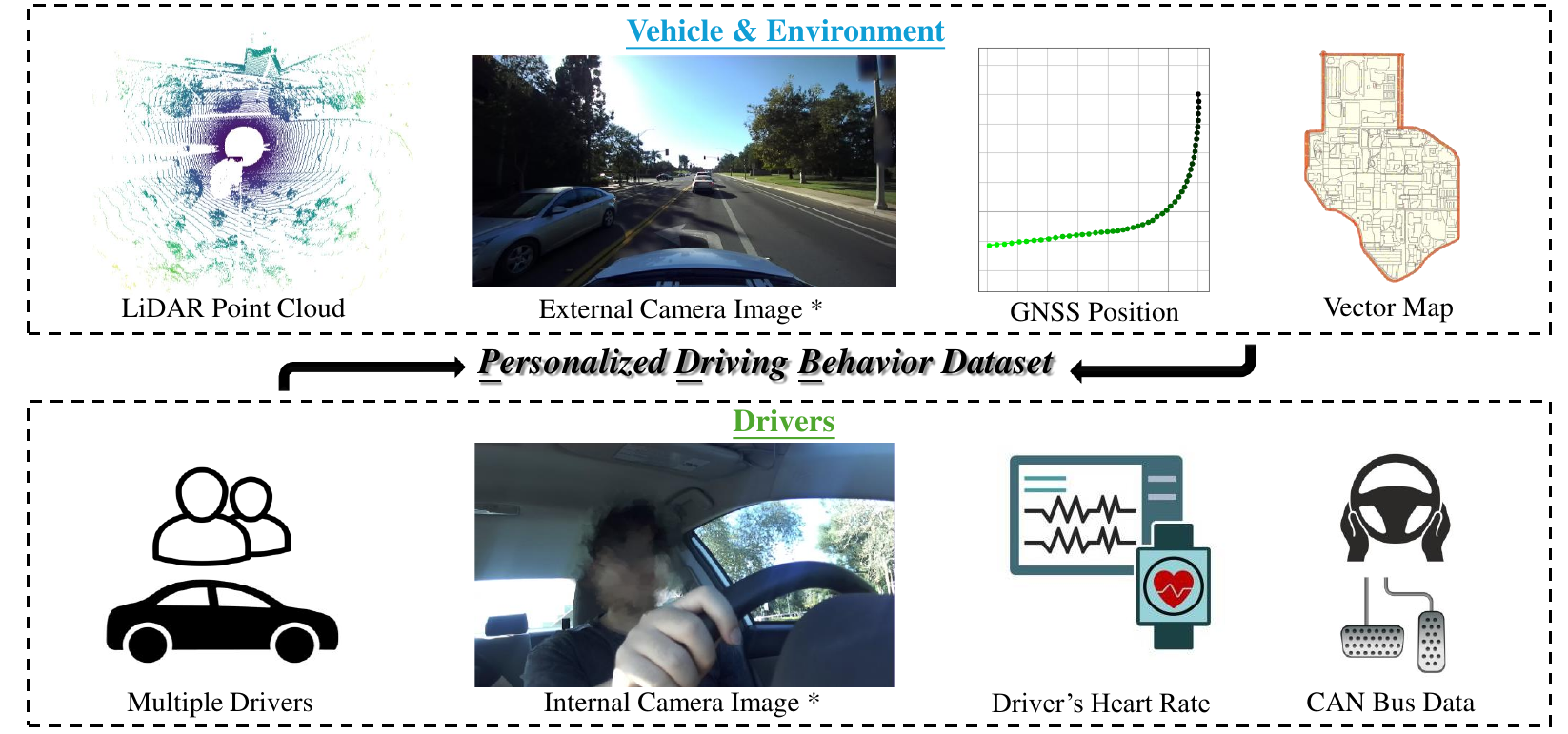}
    \caption{The overview of the PDB dataset. $^{*}$ To ensure protect privacy, details regarding driver identities have been withheld.}
    \label{fig:overview}
    \centering
  \end{figure}
}
\makeatother
\maketitle
\begin{abstract}
Driving behavior varies across individuals, yet most existing driving datasets do not preserve driver identity or control for factors that may affect driving style. We introduce the Personalized Driving Behavior (PDB) dataset, a multimodal dataset collected from 9 drivers using the same vehicle and route under comparable conditions. PDB includes data from a 128-line LiDAR, external and driver-facing cameras, GNSS, IMU, CAN-bus signals, and heart-rate measurements, with 111,060 frames of LiDAR point cloud data and more than 1.33 million images. Based on these recordings, we construct PDB-Traj, a personalized trajectory prediction benchmark with 1,309 10-second trajectory segments sampled at 10 Hz. Each segment includes the ego trajectory, surrounding-agent trajectories, and driver identity. We further introduce Gated Selective Personalization (Gate-SP), a driver-aware baseline that selectively incorporates driver identity into trajectory prediction. Experiments on PDB-Traj provide reference results for personalized motion forecasting and show that driver information can improve prediction when applied selectively. PDB and PDB-Traj provide a benchmark for studying driver-specific behavior and personalized trajectory prediction.

\end{abstract}

\section{Introduction}
\label{sec:intro}

Driving is not merely a mechanical task shaped by traffic conditions and vehicle dynamics; it is also influenced by personal preferences, cognitive responses, and habitual behaviors~\cite{liao2024review}. Even on the same road, different drivers may exhibit distinct braking, acceleration, and steering patterns because of their individual driving styles~\cite{shi2015evaluating,sysoev2017estimation}. However, most driving datasets treat drivers as interchangeable and overlook variability introduced by human decision-making. This limitation constrains the development of human-centric autonomous driving, driver behavior analysis, and adaptive assistance systems~\cite{xing2021toward}. Such personalization is especially important for trajectory prediction, where future motion depends not only on environmental context but also on driver-specific habits~\cite{li2023personalized}.

Despite growing interest in personalized driving analysis, most public datasets emphasize environmental diversity rather than driver individuality. They often anonymize driver identities or lack sufficient sensor data for driver-specific analysis, leading trajectory prediction models to assume uniform behavior. In practice, drivers may differ in acceleration, lane-changing frequency, and reaction time even under similar conditions~\cite{hang2021decision, liao2024review}. Vehicle type also affects driving behavior because different vehicles exhibit distinct acceleration, braking, and handling characteristics~\cite{wei2024ki}. Therefore, isolating driver-specific behavior requires collecting data with the same vehicle under consistent environmental conditions.

To address this gap, we introduce the \textbf{Personalized Driving Behavior (PDB) dataset}, a multimodal dataset specifically designed to study driver individuality under naturalistic yet controlled driving conditions. PDB reduces the influence of external variables by requiring all participants to drive the same vehicle along consistent routes and under comparable collection conditions. The dataset contains recordings from 9 drivers and integrates a 128-line LiDAR, external and driver-facing cameras, GNSS, IMU, CAN-bus signals, and wearable heart-rate measurements. In total, PDB includes 111,060 frames of LiDAR point cloud data and more than 1.33 million images, providing synchronized information about the surrounding environment, vehicle dynamics, and driver states.

Based on the raw PDB recordings, we further construct \textbf{PDB-Traj}, a processed benchmark for personalized trajectory prediction containing 1,309 trajectory segments, each spanning 10 seconds at 10 Hz. Each segment includes the ego-vehicle trajectory, surrounding-agent trajectories, and the corresponding driver identity, enabling both conventional and driver-specific motion forecasting under consistent vehicle and route conditions. To establish a reference for personalized prediction on this benchmark, we introduce \textbf{Gated Selective Personalization (Gate-SP)}, a driver-aware baseline that selectively incorporates driver identity into trajectory prediction. Rather than applying personalization uniformly, Gate-SP uses gated identity conditioning to introduce driver-specific information only when it is supported by the observed motion and data. Together, PDB and PDB-Traj provide complementary resources for multimodal driver behavior analysis and personalized motion prediction, while Gate-SP provides an initial benchmark for studying how driver identity can be effectively incorporated into trajectory prediction.

Our contributions are:
\begin{itemize}
\item We introduce \textbf{PDB}, a multimodal driving dataset from 9 drivers collected with the same vehicle and route while preserving driver identities.
\item We construct \textbf{PDB-Traj}, a personalized trajectory prediction benchmark with 1,309 trajectory segments and driver labels.
\item We benchmark trajectory prediction methods on PDB-Traj and introduce \textbf{Gate-SP}, a driver-aware baseline with selective identity conditioning.
\end{itemize}

\section{Related Work}
\label{sec:related_work}

\begin{table*}[!ht]
\centering
\small
\caption{A comparison of autonomous driving datasets for multi-sensor fusion.}
\begin{threeparttable} 
\begin{tabular}{lccccp{1.8cm}p{2cm}p{3cm}}
\hline
\textbf{Dataset}  & \textbf{Camera} & \textbf{LiDAR} & \textbf{GPS} & \textbf{CAN-bus} & \textbf{Driver's Behavior} & \textbf{Physiological Signal } & \textbf{Region/ Platform}\\
\hline
KITTI~\cite{geiger2012CVPR}  & $\checkmark$ & $\checkmark$& $\checkmark$ & $\times$ & $\times$ & $\times$ & Germany \\
\hline
Brain4Cars~\cite{jain2016brain4cars}  & $\checkmark$ & $\times$& $\checkmark$ & $\times$ & $\checkmark$ & $\times$ & USA \\
\hline
ApolloScape~\cite{huang2018apolloscape} &$\checkmark$ & $\checkmark$& $\checkmark$ &$\times$ & $\times$ & $\times$ & China\\
\hline
HDD~\cite{Ramanishka_2018_CVPR} &$\checkmark$ & $\checkmark$& $\checkmark$ &$\checkmark$ & $\times$ & $\times$ & USA\\
\hline
Argoverse~\cite{chang2019argoverse}  &$\checkmark$ & $\checkmark$& $\checkmark$ &$\times$ & $\times$ & $\times$ & USA\\
\hline
Lyft Level 5~\cite{lyft2019}  & $\checkmark$ & $\checkmark$& $\checkmark$ &$\times$ & $\times$ & $\times$ & USA\\
\hline
DAD ~\cite{ortega2020dmd} &$\checkmark$ & $\times$& $\times$ &$\checkmark$ & $\checkmark$ & $\checkmark$ & China\\
\hline
Waymo Open ~\cite{sun2020scalability} & $\checkmark$ & $\checkmark$& $\checkmark$ &$\times$ & $\times$ & $\times$ & USA\\
\hline
nuScenes~\cite{caesar2020nuscenes}  & $\checkmark$ & $\checkmark$& $\checkmark$ &$\checkmark$ & $\times$ & $\times$ & USA, Singapore\\
\hline
DMD~\cite{pham20203d}  & $\times$ & $\times$& $\times$ &$\times$ & $\checkmark$ & $\times$ & Spain, Simulator\\
\hline
A*3D~\cite{pham20203d}  & $\checkmark$ & $\checkmark$& $\checkmark$ &$\times$ & $\times$ & $\times$ & Singapore\\
\hline
A2D2~\cite{geyer2020a2d2}& $\checkmark$ & $\checkmark$& $\checkmark$ &$\checkmark$ & $\times$ & $\times$ & Germany\\
\hline
PandaSet~\cite{xiao2021pandaset}& $\checkmark$ & $\checkmark$& $\checkmark$ &$\times$ & $\times$ & $\times$ & USA\\
\hline
DAIR-V2X~\cite{yu2022dair}  & $\checkmark$ & $\checkmark$& $\checkmark$ &$\times$ & $\times$ & $\times$ & China \\
\hline
DriveDNA~\cite{wang2026drivedna} & $\checkmark$ & $\times$ & $\checkmark$ & $\checkmark$ & $\times$ & $\times$ & Worldwide\\

\hline
\textbf{PDB} (Ours) &$\checkmark$ & $\checkmark$& $\checkmark$ &$\checkmark$& $\checkmark$ & $\checkmark$ & USA\\
\hline
\end{tabular}
% \begin{tablenotes}
%     \item * The publication year refers to the year the dataset was introduced in the respective paper.
% \end{tablenotes}
\end{threeparttable} 
\label{dataset}
\end{table*}

Publicly available driving datasets have played a crucial role in advancing research on vehicle perception, trajectory prediction, and driver behavior modeling. However, most datasets focus on large-scale environmental diversity rather than capturing individual driver behaviors, limiting their applicability in personalized driving research~\cite{liu2024survey, wei2026beyond, wang2023multi}.

\subsection{Driver Behavior Analysis Datasets}
\label{subsec:behavior_datasets}

Understanding driver behavior is crucial for developing personalized driving models, adaptive driver assistance systems, and human-in-the-loop autonomous vehicles. Existing datasets for driver behavior analysis fall into two categories: Naturalistic Driving Study (NDS) datasets and Field Operational Test (FOT) datasets \cite{liao2024review}. NDS datasets passively collect real-world driving behavior with minimal intervention, capturing natural decision-making patterns. In contrast, FOT datasets are structured experiments designed to evaluate specific vehicle technologies under controlled conditions while still recording driver behavior.

\textbf{Naturalistic Driving Study (NDS) datasets} capture real-world driving behavior by equipping personal vehicles with sensors and cameras, ensuring high ecological validity. Representative datasets include 100-Car Naturalistic Driving Study~\cite{dingus2006100,klauer2006impact}, Candrive~\cite{marshall2013protocol}, SHRP2~\cite{blatt2015naturalistic}, MIT AVT~\cite{fridman2019advanced}, and UAH-DriveSet~\cite{romera2016need}, which collect multimodal data such as vehicle states, driver maneuvers, heart rate, gaze tracking, and facial expressions \cite{blatt2015naturalistic, fridman2019advanced, romera2016need}. Additional datasets, such as Brain4Cars~\cite{jain2016brain4cars}, Dr(eye)ve~\cite{palazzi2018predicting}, Drive\&Act~\cite{martin2019drive}, and DMD~\cite{ortega2020dmd}, extend NDS research by integrating driver attention and physiological monitoring. However, due to varying vehicle types, road conditions, and lack of experimental control, isolating driver-specific behaviors remains a challenge. Furthermore, anonymized driver identities hinder personalized driving style analysis. More recently, DriveDNA~\cite{wang2026drivedna} provides a large-scale naturalistic dataset for personalized driving behavior analysis with persistent driver identities.

\textbf{Field Operational Test (FOT) datasets} are designed to systematically evaluate how drivers interact with specific vehicle technologies, such as lane-keeping assist, adaptive cruise control, and collision warning systems. Unlike NDS datasets, FOT data is collected under controlled scenarios with predefined evaluation criteria. Studies such as those by Viti et al.\cite{viti2008driving} and Lyu et al. \cite{lyu2022using} examine how drivers adapt to intelligent vehicle systems through small-scale, naturalistic FOT (N-FOT) studies, where participants are monitored while driving under predefined conditions. Larger-scale projects, such as those managed by the Dutch Ministry of Transport~\cite{alkim2007field}, have explored driver responses to automation technologies and vehicle-based interventions. While FOT datasets offer more control over experimental conditions than NDS datasets, their primary focus is on evaluating vehicle functions rather than understanding individual driver behavior. Additionally, they often involve a limited number of drivers and controlled exposure to specific technologies, limiting their applicability for studying long-term, naturally occurring driving patterns.

Despite the contributions of both NDS and FOT datasets, they lack systematic vehicle control, leading to variations in acceleration, braking, and steering due to differing vehicle dynamics rather than driver behavior. Additionally, most datasets do not integrate structured trajectory data, making it difficult to link driving style with motion patterns. The Personalized Driving Behavior (PDB) dataset addresses these limitations by collecting multi-modal driving data under controlled conditions, ensuring all participants drive the same vehicle on identical routes to eliminate confounding factors.

\begin{figure*}[t]
\centering
\includegraphics[width=0.85\linewidth]{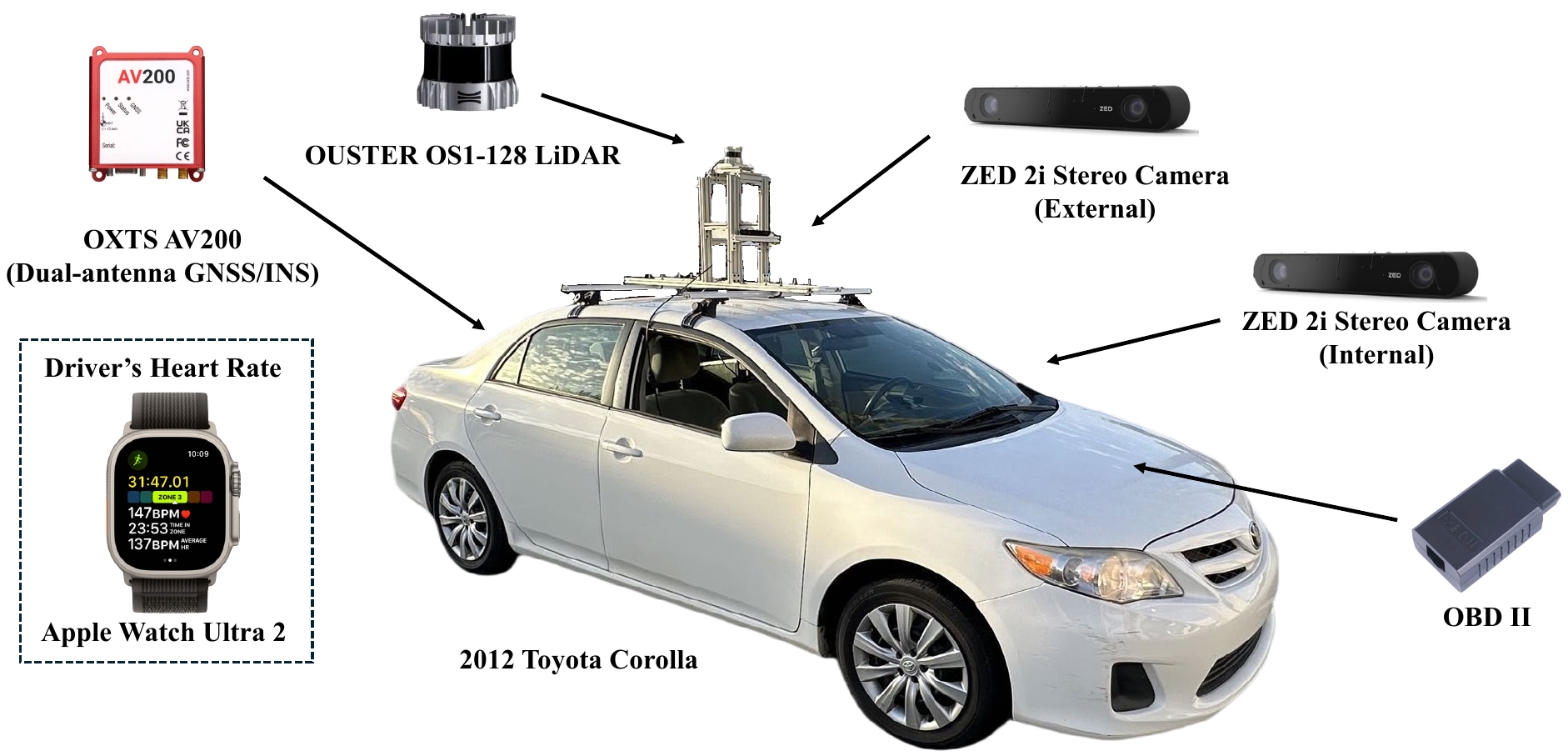}
\caption{The PDB data collection sensor setup.}
\label{Fig:vehicle_setup}
\end{figure*}

\subsection{Vehicle Trajectory Prediction Datasets}
\label{subsec:trajectory_datasets}

Early trajectory prediction datasets primarily focused on large-scale vehicle motion data without considering driver-specific behaviors. NGSIM~\cite{alexiadis2004next} provides detailed vehicle trajectory information widely used for traffic flow analysis and modeling. Similarly, HighD \cite{krajewski2018highd} offers high-precision highway trajectory data, supporting lane-changing behavior and traffic dynamics studies. Other datasets with similar objectives include inD~\cite{bock2020ind}, which specializes in intersection scenarios, and SinD~\cite{xu2022drone}, which incorporates traffic signal information at intersections. While valuable for understanding general vehicle dynamics, these datasets lack multi-modal sensor inputs and do not include driver identity information, making them unsuitable for personalized trajectory prediction.

To enhance motion forecasting in complex urban scenarios, later datasets integrated high-resolution sensor data, including LiDAR, cameras, and radar. nuScenes \cite{caesar2020nuscenes} and Waymo Open Dataset \cite{sun2020scalability} provide extensive multi-modal data, advancing perception-driven trajectory modeling. Additional datasets, such as Argoverse~\cite{chang2019argoverse,wilson2023argoverse}, Lyft Level 5~\cite{houston2021one}, and PandaSet~\cite{xiao2021pandaset}, offer diverse urban driving trajectories and sensor-rich environments. These datasets have significantly contributed to trajectory prediction research but anonymize driver identities and focus primarily on vehicle-centric motion rather than individual driving behaviors. Consequently, most trajectory prediction models trained on these datasets assume uniform driver behavior, limiting their adaptability to personalized motion forecasting.

Despite advances in trajectory prediction, existing datasets do not account for driver-specific variability, making it difficult to model individual driving tendencies and decision-making patterns. Addressing this limitation requires datasets that integrate vehicle trajectories with driver identity information to enable personalized trajectory prediction and adaptive driving assistance.

\subsection{Comprehensive Multi-Modal Driving Datasets}

Many large-scale datasets provide extensive sensory data for autonomous driving research. For example, KITTI~\cite{geiger2012CVPR}, ApolloScape~\cite{huang2018apolloscape}, and DAIR-V2X~\cite{yu2022dair} offer high-resolution data from LiDAR, stereo cameras, and GPS. Additionally, datasets such as DAIR-V2X~\cite{yu2022dair} provide multi-agent cooperative perception data. These datasets have significantly contributed to advancements in object detection, semantic segmentation, and vehicle-to-everything (V2V) communication. However, their primary focus remains on perception-based tasks rather than driver behavior modeling.

Currently, only a few datasets such as HDD~\cite{Ramanishka_2018_CVPR}, DAD dataset~\cite{Qiu2019driving}, and A2D2~\cite{geyer2020a2d2} include CAN-bus data, which facilitates the identification of driver maneuvers and enables the monitoring of various vehicle states. Moreover, datasets like DMD~\cite{ortega2020dmd} provides valuable insights into driver monitoring but lack structured trajectory data, limiting their applicability in personalized motion forecasting. While the DAD dataset offers information on driver states through in-cabin cameras and chest band sensors, along with CAN-bus data, it lacks GPS trajectory information~\cite{Qiu2019driving}. Therefore, there remains a need for a more comprehensive dataset that integrates environmental perception data with both driver and vehicle state information in open-road driving scenarios. Such a dataset would be instrumental in advancing research on driver decision-making processes.

\begin{figure}[t]
\centering
\includegraphics[width=0.95\linewidth]{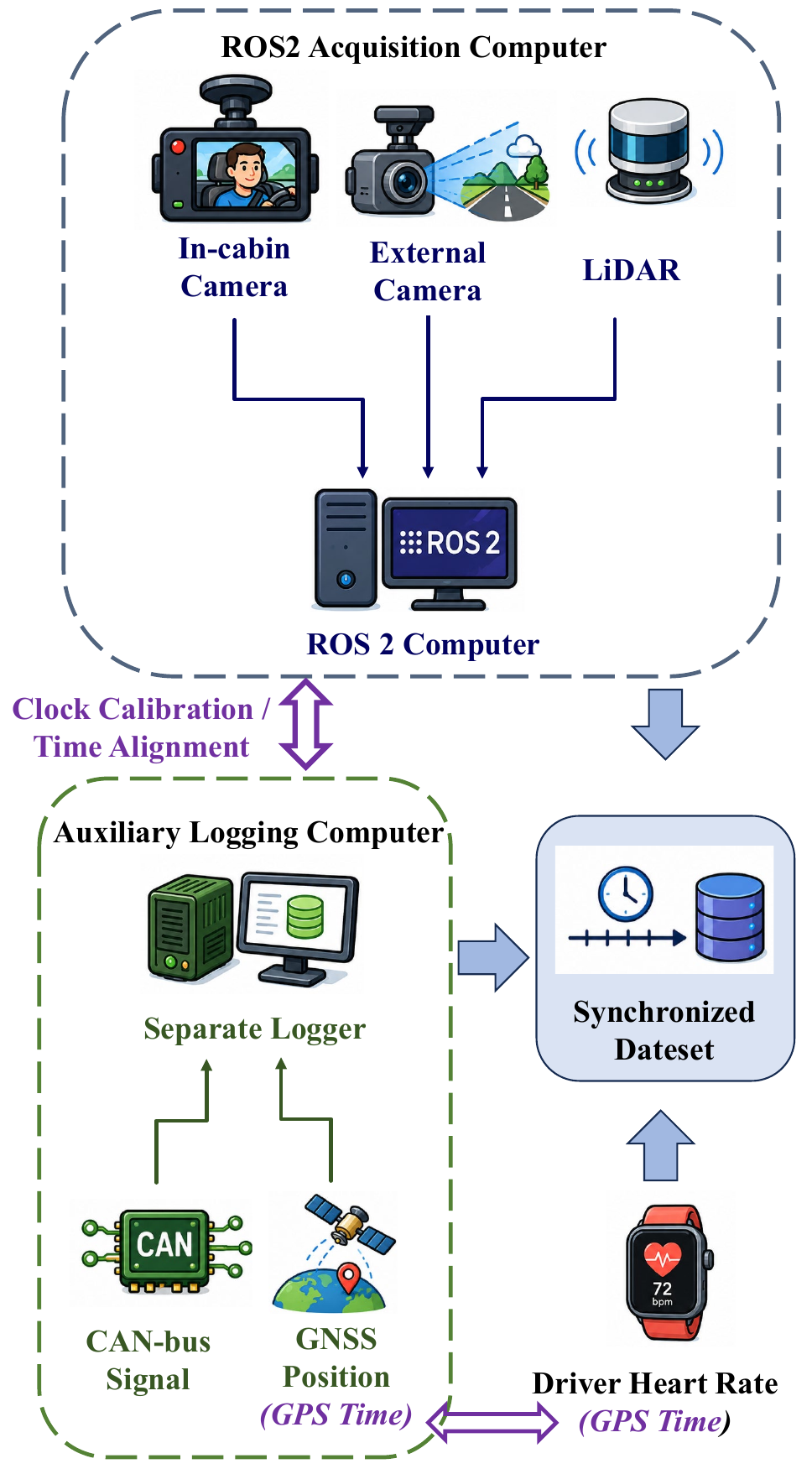}
\caption{Time synchronization of the multimodal PDB data streams. 
GNSS provides the GPS time reference for the auxiliary logger, whose clock is used to synchronize the ROS~2 acquisition computer. Driver heart-rate measurements are independently GPS-timestamped and aligned to the common timeline during post-processing.}
\label{fig:time_sync}
\end{figure}

\section{Dataset Design and Collection Methodology}
\subsection{Dataset Overview}

The PDB dataset is designed to capture multimodal driving data for studying individual driving behavior under a consistent collection protocol. Because driving style, reaction time, and decision-making can vary substantially across individuals, PDB preserves driver identities and records both external traffic conditions and driver-related measurements.

The dataset contains data from 9 drivers collected across multiple driving sessions along a predefined urban route. Each session provides synchronized measurements from multiple sensing modalities, enabling the joint analysis of the surrounding environment, vehicle dynamics, and driver states. In total, PDB contains 111,060 LiDAR point clouds and more than 1.33 million images. Based on these recordings, we further construct \textbf{PDB-Traj}, a trajectory prediction benchmark containing 1,309 segments. Each segment spans 10 seconds and is sampled at 10 Hz, while adjacent segments have a 5-second temporal overlap.

\subsection{Sensor Setup}

The test vehicle, a 2012 Toyota Corolla, is equipped with an integrated sensor suite for recording the surrounding environment, vehicle dynamics, and driver behavior. The sensor specifications and installation configuration are illustrated in \textbf{Figure~\ref{Fig:vehicle_setup}} and summarized as follows:

\begin{itemize}
\item \textbf{OUSTER OS1-128 LiDAR:} The LiDAR is mounted on a stable roof-rack structure and captures high-resolution 3D point clouds at 10 Hz. It provides spatial measurements of surrounding vehicles, pedestrians, cyclists, static objects, and road infrastructure.

\item \textbf{ZED 2i Stereo Cameras:} Two ZED 2i stereo cameras are installed on the vehicle. The external camera is mounted on the roof and faces forward to capture road scenes and traffic interactions. The internal camera faces the driver and records facial appearance, head movement, gaze direction, and other visual indicators related to driver attention and behavior.

\item \textbf{OXTS AV200 GNSS Unit:} The GNSS unit records the ego vehicle's geographic position, including latitude, longitude, and altitude, together with the motion information required for ego-trajectory construction. These measurements provide the global reference used to align vehicle motion with the recorded sensor data.

\item \textbf{OBD-II Interface:} The OBD-II interface accesses the vehicle's Controller Area Network (CAN bus) and records operational signals such as vehicle speed, throttle position, brake pressure, and steering-wheel angle. These signals provide direct measurements of driver control actions and vehicle dynamics.

\item \textbf{Apple Watch Ultra 2:} Each driver wore an Apple Watch Ultra 2 during data collection to record physiological measurements, including heart rate. These measurements support the analysis of potential relationships between physiological responses and driving behavior.

\end{itemize}

LiDAR point clouds, camera images, and IMU measurements were recorded through the Robot Operating System 2 (ROS~2) and stored in structured ROS~2 bag files. GNSS measurements were recorded concurrently using the OXTS AV200 unit, while CAN-bus signals were collected using a custom-developed interface. Measurements from the different sensing modalities were timestamped and synchronized using a common time reference. Heart-rate measurements were recorded separately and temporally aligned with the vehicle and environmental sensor data.

\subsection{Participants and Driving Scenarios}

The driving sessions were conducted on a university campus in University of California Riverside, CA, USA, following a fixed-loop route to maintain consistency across participants. Data were collected during comparable weekday periods, capturing naturalistic urban traffic interactions at intersections, pedestrian crossings, and mixed-traffic areas. Each driving session lasted approximately 20 minutes, providing sufficient exposure to different traffic conditions and driving maneuvers. The same vehicle and route were used throughout the study, while collection time and weather conditions were kept comparable whenever possible.

A total of 9 participants were recruited, with each participant completing multiple driving sessions. This repeated-session design captures both differences across drivers and variations in the behavior of the same driver across separate trips. By preserving driver identities and maintaining a consistent vehicle and route, the dataset supports both general driving behavior analysis and personalized trajectory prediction.

The experimental protocol primarily controlled the vehicle platform and driving route while retaining natural variations in surrounding traffic. This design reduces major confounding factors caused by vehicle dynamics and route differences without removing the realistic traffic interactions needed to study natural driving behavior.

\begin{table*}[ht!]
\centering
\caption{Sensor specifications and sampling frequencies.}
\label{tab:sensors}
\small
\renewcommand{\arraystretch}{1.15}
\begin{threeparttable}
\begin{tabular}{ccc}
\hline
\textbf{Data Type} & \textbf{Sensor Model} & \textbf{Frequency (Hz)} \\
\hline
Point Cloud
& OUSTER OS1-128 LiDAR
& 10 \\
\hline
\begin{tabular}[c]{@{}c@{}}
Raw Images \\
(External and Internal Cameras)
\end{tabular}
& ZED 2i Stereo Cameras
& 30 \\
\hline

GNSS Position
& OXTS AV200 GNSS Unit
& 100 \\
\hline

CAN-Bus Signals
& OBD-II Interface
&
\begin{tabular}[c]{@{}c@{}}
SPEED: 83.33$^{\dagger}$ \qquad STEER\_ANGLE: 83.33$^{\dagger}$ \\
GAS\_PEDAL: 31.25 $^{\dagger}$ \qquad BRAKE\_PRESSURE: 41.67 $^{\dagger}$
\end{tabular} \\
\hline

IMU
& ZED 2i Stereo Camera
& 400 \\
\hline

Driver Heart Rate
& Apple Watch Ultra 2
& 0.2 on average \\
\hline
\end{tabular}

\begin{tablenotes}
\small
\item[$\dagger$] CAN-bus signals have different sampling frequencies because they are recorded through an event-driven callback mechanism. Four representative signals are reported.
\end{tablenotes}
\end{threeparttable}
\end{table*}

\section{Data Processing and Calibration}
\label{sec:data_processing}

\subsection{Sensor Synchronization}

Accurate sensor synchronization is crucial for multimodal data collection, particularly for trajectory estimation, behavior analysis, and sensor fusion~\cite{wei2025integrating}. Timestamp misalignment across sensors can introduce inconsistencies in motion tracking and reduce the reliability of downstream analyses. Figure~\ref{fig:time_sync} illustrates the synchronization scheme used in PDB.

The external and in-cabin cameras, LiDAR, and camera IMU were acquired through ROS 2~\cite{macenski2022robot} on the primary acquisition computer. GNSS and CAN-bus measurements were recorded on a separate logging computer. The GNSS receiver supplied a GPS-referenced time source to the logging system, and the ROS 2 computer was synchronized accordingly, providing a common temporal reference across the two acquisition platforms.

Driver heart-rate measurements were recorded independently using the Apple Watch with GPS-referenced timestamps. Owing to the substantially lower sampling rate of the heart-rate signal, these measurements were associated with the nearest synchronized vehicle timestamp during post-processing.

All sensor streams retain their original timestamps, while the processed dataset is aligned to the shared temporal reference. Since GNSS measurements and LiDAR frames are acquired asynchronously, the ego-vehicle position at each LiDAR timestamp is obtained by linear interpolation between the temporally adjacent GNSS measurements.

To derive the ego-vehicle position at a LiDAR timestamp, we apply linear interpolation between the two temporally adjacent GNSS measurements:
\begin{equation}
P_t = P_{t_1} + \frac{t-t_1}{t_2-t_1}\left(P_{t_2}-P_{t_1}\right),
\label{eq:gnss_interpolation}
\end{equation}
where \(P_t\) is the estimated vehicle position at LiDAR timestamp \(t\), \(P_{t_1}\) and \(P_{t_2}\) are the two nearest GNSS readings before and after \(t\), and \(t_1, t_2\) are their respective timestamps. This interpolation strategy ensures that the vehicle trajectory is smoothly aligned with the high-frequency LiDAR frames, enabling precise motion estimation.

\begin{figure*}
\centering
\includegraphics[width=\linewidth]{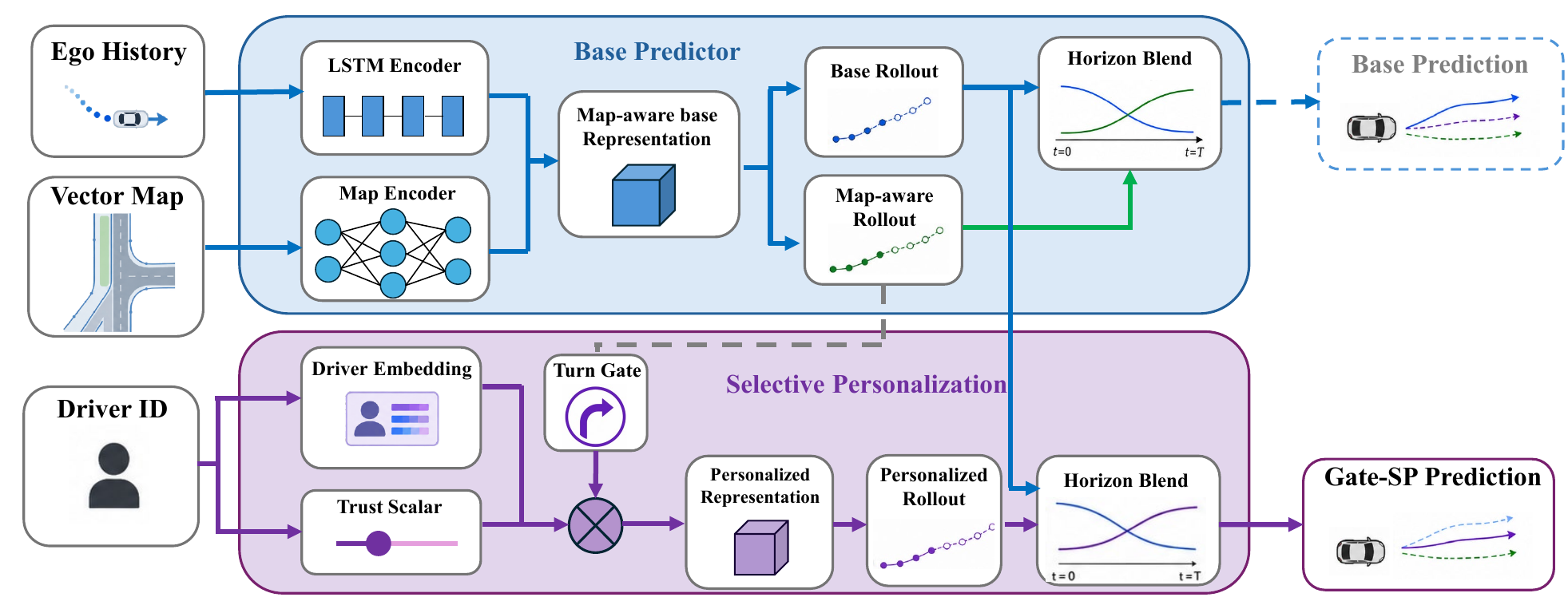}
\caption{Overview of Gate-SP. Ego history and vector-map context are first encoded to produce base and map-aware rollouts, which are fused through a horizon blend for base prediction. Driver-specific information is then selectively incorporated through a turn gate and a trust scalar to form a personalized rollout, which is blended with the base rollout to produce the final Gate-SP prediction.}
\label{fig:Gate-SP}
\end{figure*}

\subsection{Sensor Calibration}

\paragraph{Camera Calibration}
Camera calibration determines the intrinsic parameters of each camera, including focal length, principal point, and distortion coefficients. We use the checkerboard-based Zhang calibration method~\cite{zhang2002flexible} to estimate these parameters and correct image distortion.

\paragraph{Camera-LiDAR Extrinsic Calibration}
We estimate the extrinsic transformation between the external camera and LiDAR using a target-based calibration procedure. Correspondences between image features and 3D LiDAR points are used to minimize reprojection error. The resulting calibration supports LiDAR-to-image projection and other multimodal applications, while PDB-Traj focuses on trajectory modeling using the aligned motion data.

\section{PDB-Traj Benchmark}
\label{sec:benchmark}

\subsection{PDB-Traj}
\label{subsec:pdbtraj}

PDB-Traj is constructed through a post-processing pipeline that converts the raw PDB sensor recordings into structured multi-agent trajectory sequences. The pipeline begins with the LiDAR point clouds, where a PointPillars detector~\cite{lang2019pointpillars} is applied to identify surrounding vehicles, bicycles, and pedestrians. The resulting 3D detections are associated across consecutive frames using AB3DMOT~\cite{weng2020ab3dmot}. The generated tracks are subsequently inspected and refined to correct evident detection and association errors.

The ego-vehicle GNSS measurements are first processed using Kalman filtering~\cite{kalman1960new} to reduce positioning noise. The filtered positions are then interpolated at the LiDAR timestamps using Equation~\eqref{eq:gnss_interpolation}, producing a temporally aligned ego trajectory. Each tracked 3D object is first represented in the ego-vehicle coordinate system and then transformed into the global coordinate system using the corresponding ego position and heading, which produces temporally aligned trajectories for both the ego vehicle and surrounding agents. The route map is derived from OpenStreetMap and manually simplified to retain the road elements relevant to the collected route, providing lane geometry, intersections, and pedestrian crossings as spatial context for the trajectories.

The final benchmark contains 1{,}309 segments. Each segment spans 10 seconds with 100 frames sampled at 10\,Hz, and consecutive segments are generated with a 5-second temporal overlap. Each segment includes the ego-vehicle trajectory, surrounding-agent trajectories, and the associated driver identity.

The benchmark task is to predict the future ego positions over a 3\,s or 5\,s horizon from a 3\,s observation, at 10\,Hz. Each segment yields two windows, giving 2{,}618 samples per configuration from the 9 drivers. We use a clip-level 80/20 train/test split, stratified by driver, since windows from the same clip overlap temporally and a sample-level split would leak near-identical trajectories across partitions. We report Average and Final Displacement Error (ADE/FDE) in meters. Because straight segments account for most of the samples, we additionally report a balanced FDE (balFDE), the mean of the straight-stratum and turn-stratum FDE, which was fixed before evaluation. Multimodal methods are scored by their most-likely mode; full protocol details and best-of-$K$ metrics appear in the supplementary material.

\begin{figure*}[t]
\centering
\includegraphics[width=\linewidth]{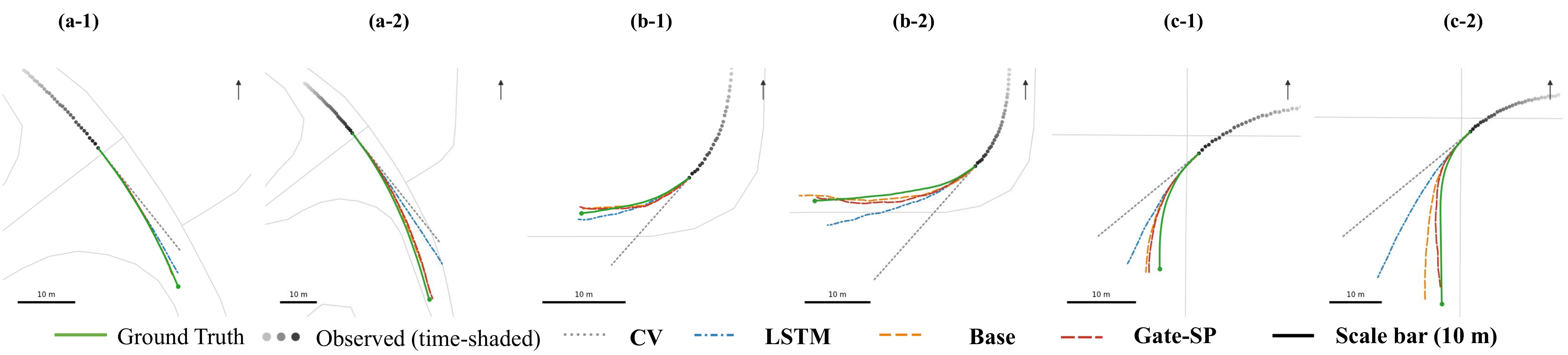}
\caption{Qualitative trajectory prediction results on PDB-Traj. Three representative cases from different drivers are shown. Panels (a-1), (b-1), and (c-1) use 3 s observation and 3 s prediction, while (a-2), (b-2), and (c-2) use 3 s observation and 5 s prediction. Gray points denote the observed trajectory, and the green curve denotes the ground truth. CV, LSTM, Base, and Gate-SP predictions are shown for comparison.}

\label{fig:qualitative_results}
\end{figure*}

\subsection{Gate-SP}
\label{subsec:gatesp}

We benchmark six reference predictors: a constant velocity model (CV), a constant turn rate and velocity model (CTRV)~\cite{schubert2008comparison}, an agent-centric LSTM~\cite{graves2012long} encoder-decoder that follows the trajectory processing of Social-LSTM~\cite{alahi2016social}, a re-implementation of Trajectron++~\cite{salzmann2020trajectron} with interaction encoding and a CVAE, a compact Transformer~\cite{vaswani2017attention} (TF) with masked neighbor attention and winner-takes-all mode queries, and a Transformer with driver embedding (TF-DE), which augments TF with a learnable per-driver embedding and serves as a plain identity-conditioning reference. All learned methods share the same data loading, split, and evaluation code. On top of this suite, we introduce Gated Selective Personalization (Gate-SP), a driver-aware baseline that tests whether the preserved driver identities improve ego-trajectory prediction. \textit{Further implementation details of Gate-SP are provided in Appendix~A, while additional experimental results are reported in Appendix~B.}

Figure~\ref{fig:Gate-SP} illustrates Gate-SP. The base predictor encodes the agent-centric ego observation with a two-layer LSTM and conditions on a 22-dimensional road-context vector derived from the OSM map, covering the nearest-road offset and heading, the distance to the next intersection, its branch directions, and signal and crossing flags. The road-context representation is introduced through a zero-initialized map correction and a horizon-dependent blend: the early prediction follows the map-free rollout and smoothly transitions to the map-corrected rollout at longer lead times, where road geometry increasingly constrains the path.

Preliminary experiments showed that naive identity conditioning fails to improve this base. Straights dominate the data, so an always-on per-driver perturbation regresses them, and per-driver parameters overfit the drivers whose headings are derived from position-only GNSS and are therefore noisier in sharp maneuvers. Gate-SP makes personalization selective through the correction
\begin{equation}
\Delta_{p} = g(\tau)\cdot s_{d}\cdot \phi(\mathbf{e}_{d}),
\label{eq:pterm}
\end{equation}
where $\mathbf{e}_{d}\in\mathbb{R}^{16}$ is a learned embedding of driver $d$, $\phi$ is a zero-initialized projection, $g(\tau)$ is a turn self-gate driven by the base model's own predicted heading change $\tau$, and $s_{d}$ is a per-driver trust scalar initialized near zero. The turn gate suppresses personalization on straight segments, and the trust scalar limits personalization for drivers whose data do not support a reliable driver-specific correction. Since $\phi$ is zero-initialized and regularized toward zero, Gate-SP matches the base at initialization and is encouraged to introduce driver-specific corrections only when they are supported by the training data.

\begin{table}[t]
\centering
\footnotesize
\setlength{\tabcolsep}{3.2pt}
\caption{Benchmark results on the PDB-Traj test split (meters, lower is better). Multimodal methods are scored by their most-likely mode. Base is the map-conditioned predictor without personalization; Gate-SP adds the selective identity conditioning.}
\label{tab:main_results}
\begin{tabular}{lcccccc}
\hline
 & \multicolumn{3}{c}{3s$\rightarrow$3s} & \multicolumn{3}{c}{3s$\rightarrow$5s} \\
Method & ADE & FDE & balFDE & ADE & FDE & balFDE \\
\hline
CV~\cite{schubert2008comparison} & 1.59 & 4.05 & 5.17 & 3.91 & 10.23 & 11.62 \\
CTRV~\cite{schubert2008comparison} & 1.36 & 3.53 & 3.55 & 3.46 & 9.40 & 9.48 \\
LSTM~\cite{graves2012long,alahi2016social} & 1.00 & 2.55 & 2.79 & 2.53 & 6.85 & 7.29 \\
Transformer (TF)~\cite{vaswani2017attention} & 2.43 & 5.08 & 5.69 & 4.73 & 11.13 & 11.99 \\
TF-DE & 1.81 & 4.28 & 4.58 & 4.01 & 9.82 & 10.24 \\
Trajectron++~\cite{salzmann2020trajectron} & 1.46 & 3.75 & 4.99 & 4.07 & 10.32 & 11.65 \\
\hline
Base (w/o personalization) & 0.89 & 2.15 & 2.19 & 1.96 & 5.01 & 5.12 \\
Gate-SP & 0.88 & 2.15 & 2.19 & 1.91 & 4.85 & 4.98 \\
\hline
\end{tabular}
\end{table}

\paragraph{Quantitative Analysis.} Table~\ref{tab:main_results} summarizes the benchmark. Learned dynamics provide the largest single improvement, reducing FDE at 5\,s from 10.23 (CV) to 6.85 (LSTM). Road context provides the second, reducing it further to 5.01, and the stratified results in the supplementary material show that this gain concentrates on turning and hard-maneuver samples. Gate-SP further reduces the 5\,s FDE to 4.85, indicating that selective personalization provides an additional benefit beyond the map-conditioned base. The interaction-aware and attention-based baselines fall behind the LSTM, suggesting that these higher-capacity architectures may be less effective at the current dataset scale.

\paragraph{Qualitative Analysis.}
Figure~\ref{fig:qualitative_results} shows representative predictions from the PDB-Traj test set. On relatively simple motion patterns, the learned predictors generally follow the ground-truth trajectory closely. The differences become more evident during turning maneuvers, where the CV baseline tends to deviate substantially as the prediction horizon increases. Incorporating road context allows the Base model to better follow the route geometry, while Gate-SP further adjusts the prediction in cases where driver-specific behavior is informative. These examples are consistent with the quantitative results, where the benefit of personalization is more pronounced at the longer prediction horizon.

\section{Conclusion}
\label{sec:conclusion}

This paper introduced PDB, a multimodal dataset for studying personalized driving behavior, and PDB-Traj, a trajectory prediction benchmark derived from it. By using the same vehicle and route while preserving driver identities, PDB reduces major confounding factors associated with vehicle dynamics and route differences. Benchmark experiments establish reference performance across physics-based, learned, and map-aware predictors, and show that road context is particularly useful for turning maneuvers where motion history alone provides limited guidance.

The experiments further show that naive identity conditioning does not improve a strong baseline, while selective personalization is more effective for longer-horizon prediction. Gate-SP uses a turn self-gate and a driver-specific trust scalar to control when identity information is introduced, improving the base model at the 5\,s horizon. PDB and PDB-Traj provide a foundation for further research on personalized trajectory prediction, driver behavior modeling, and human-centric driving systems.

{
    \small
    \bibliographystyle{ieeenat_fullname}
    \bibliography{main}

@String(CVPR= {IEEE Conf. Comput. Vis. Pattern Recog.})

@String(CVPRW= {IEEE Conf. Comput. Vis. Pattern Recog. Worksh.})

@String(CVPR  = {CVPR})

@String(CVPRW= {CVPRW})

@article{liao2024review,
  title={A review of personalization in driving behavior: Dataset, modeling, and validation},
  author={Liao, Xishun and Zhao, Zhouqiao and Barth, Matthew J and Abdelraouf, Amr and Gupta, Rohit and Han, Kyungtae and Ma, Jiaqi and Wu, Guoyuan},
  journal={IEEE Transactions on Intelligent Vehicles},
  year={2024},
  publisher={IEEE}
}

@inproceedings{wei2024ki,
  title={KI-GAN: Knowledge-informed generative adversarial networks for enhanced multi-vehicle trajectory forecasting at signalized intersections},
  author={Wei, Chuheng and Wu, Guoyuan and Barth, Matthew J and Abdelraouf, Amr and Gupta, Rohit and Han, Kyungtae},
  booktitle={2024 IEEE/CVF Conference on Computer Vision and Pattern Recognition Workshops (CVPRW)},
  pages={7115--7124},
  year={2024},
  organization={IEEE}
}

@article{shi2015evaluating,
  title={Evaluating driving styles by normalizing driving behavior based on personalized driver modeling},
  author={Shi, Bin and Xu, Li and Hu, Jie and Tang, Yun and Jiang, Hong and Meng, Wuqiang and Liu, Hui},
  journal={IEEE Transactions on Systems, Man, and Cybernetics: Systems},
  volume={45},
  number={12},
  pages={1502--1508},
  year={2015},
  publisher={IEEE}
}

@article{sysoev2017estimation,
  title={Estimation of the driving style based on the users’ activity and environment influence},
  author={Sysoev, Mikhail and Kos, Andrej and Guna, Jo{\v{z}}e and Poga{\v{c}}nik, Matev{\v{z}}},
  journal={Sensors},
  volume={17},
  number={10},
  pages={2404},
  year={2017},
  publisher={MDPI}
}

@article{xing2021toward,
  title={Toward human-vehicle collaboration: Review and perspectives on human-centered collaborative automated driving},
  author={Xing, Yang and Lv, Chen and Cao, Dongpu and Hang, Peng},
  journal={Transportation research part C: emerging technologies},
  volume={128},
  pages={103199},
  year={2021},
  publisher={Elsevier}
}

@inproceedings{li2023personalized,
  title={Personalized trajectory prediction for driving behavior modeling in ramp-merging scenarios},
  author={Li, Siyan and Wei, Chuheng and Wu, Guoyuan and Barth, Matthew J and Abdelraouf, Amr and Gupta, Rohit and Han, Kyungtae},
  booktitle={2023 Seventh IEEE International Conference on Robotic Computing (IRC)},
  pages={1--4},
  year={2023},
  organization={IEEE}
}

@article{hang2021decision,
  title={Decision making of connected automated vehicles at an unsignalized roundabout considering personalized driving behaviours},
  author={Hang, Peng and Huang, Chao and Hu, Zhongxu and Xing, Yang and Lv, Chen},
  journal={IEEE Transactions on Vehicular Technology},
  volume={70},
  number={5},
  pages={4051--4064},
  year={2021},
  publisher={IEEE}
}

@techreport{klauer2006impact,
  title={The impact of driver inattention on near-crash/crash risk: An analysis using the 100-car naturalistic driving study data},
  author={Klauer, Sheila G and Dingus, Thomas A and Neale, Vicki L and Sudweeks, Jeremy D and Ramsey, David J},
  year={2006},
  institution={United States. Department of Transportation. National Highway Traffic Safety~…}
}

@techreport{dingus2006100,
  title={The 100-car naturalistic driving study, Phase II-results of the 100-car field experiment},
  author={Dingus, Thomas A and Klauer, Sheila G and Neale, Vicki Lewis and Petersen, Andy and Lee, Suzanne E and Sudweeks, Jeremy and Perez, Miguel A and Hankey, Jonathan and Ramsey, David and Gupta, Santosh and others},
  year={2006},
  institution={United States. Department of Transportation. National Highway Traffic Safety~…}
}

@article{marshall2013protocol,
  title={Protocol for Candrive II/Ozcandrive, a multicentre prospective older driver cohort study},
  author={Marshall, Shawn C and Man-Son-Hing, Malcolm and Bedard, Michel and Charlton, Judith and Gagnon, Sylvain and Gelinas, Isabelle and Koppel, Sjaan and Korner-Bitensky, Nicol and Langford, Jim and Mazer, Barbara and others},
  journal={Accident Analysis \& Prevention},
  volume={61},
  pages={245--252},
  year={2013},
  publisher={Elsevier}
}

@techreport{blatt2015naturalistic,
  title={Naturalistic driving study: Field data collection},
  author={Blatt, Alan and Pierowicz, John and Flanigan, Marie and Lin, Pei-Sung and Kourtellis, Achilleas and Lee, Chanyoung and Jovanis, Paul and Jenness, James and Wilaby, Martha and Campbell, John and others},
  year={2015},
  institution = {National Academies of Sciences, Engineering, and Medicine}
}

@article{fridman2019advanced,
  title={MIT advanced vehicle technology study: Large-scale naturalistic driving study of driver behavior and interaction with automation},
  author={Fridman, Lex and Brown, Daniel E and Glazer, Michael and Angell, William and Dodd, Spencer and Jenik, Benedikt and Terwilliger, Jack and Patsekin, Aleksandr and Kindelsberger, Julia and Ding, Li and others},
  journal={IEEE Access},
  volume={7},
  pages={102021--102038},
  year={2019},
  publisher={IEEE}
}

@inproceedings{romera2016need,
  title={Need data for driver behaviour analysis? Presenting the public UAH-DriveSet},
  author={Romera, Eduardo and Bergasa, Luis M and Arroyo, Roberto},
  booktitle={2016 IEEE 19th international conference on intelligent transportation systems (ITSC)},
  pages={387--392},
  year={2016},
  organization={IEEE}
}

@article{jain2016brain4cars,
  title={Brain4cars: Car that knows before you do via sensory-fusion deep learning architecture},
  author={Jain, Ashesh and Koppula, Hema S and Soh, Shane and Raghavan, Bharad and Singh, Avi and Saxena, Ashutosh},
  journal={arXiv preprint arXiv:1601.00740},
  year={2016}
}

@article{palazzi2018predicting,
  title={Predicting the Driver's Focus of Attention: the DR (eye) VE Project},
  author={Palazzi, Andrea and Abati, Davide and Solera, Francesco and Cucchiara, Rita and others},
  journal={IEEE transactions on pattern analysis and machine intelligence},
  volume={41},
  number={7},
  pages={1720--1733},
  year={2018},
  publisher={IEEE}
}

@inproceedings{martin2019drive,
  title={Drive\&act: A multi-modal dataset for fine-grained driver behavior recognition in autonomous vehicles},
  author={Martin, Manuel and Roitberg, Alina and Haurilet, Monica and Horne, Matthias and Rei{\ss}, Simon and Voit, Michael and Stiefelhagen, Rainer},
  booktitle={Proceedings of the IEEE/CVF International Conference on Computer Vision},
  pages={2801--2810},
  year={2019}
}

@inproceedings{ortega2020dmd,
  title={Dmd: A large-scale multi-modal driver monitoring dataset for attention and alertness analysis},
  author={Ortega, Juan Diego and Kose, Neslihan and Ca{\~n}as, Paola and Chao, Min-An and Unnervik, Alexander and Nieto, Marcos and Otaegui, Oihana and Salgado, Luis},
  booktitle={Computer Vision--ECCV 2020 Workshops: Glasgow, UK, August 23--28, 2020, Proceedings, Part IV 16},
  pages={387--405},
  year={2020},
  organization={Springer}
}

@inproceedings{viti2008driving,
  title={Driving behavior interaction with ACC: results from a Field Operational Test in the Netherlands},
  author={Viti, Francesco and Hoogendoorn, Serge P and Alkim, Tom P and Bootsma, Gerben},
  booktitle={2008 IEEE Intelligent Vehicles Symposium},
  pages={745--750},
  year={2008},
  organization={IEEE}
}

@article{lyu2022using,
  title={Using naturalistic driving data to identify driving style based on longitudinal driving operation conditions},
  author={Lyu, Nengchao and Wang, Yugang and Wu, Chaozhong and Peng, Lingfeng and Thomas, Alieu Freddie},
  journal={Journal of intelligent and connected vehicles},
  volume={5},
  number={1},
  pages={17--35},
  year={2022},
  publisher={TUP}
}

@inproceedings{alkim2007field,
  title={Field operational test" the assisted driver"},
  author={Alkim, Tom P and Bootsma, Gerben and Hoogendoorn, Serge P},
  booktitle={2007 IEEE Intelligent Vehicles Symposium},
  pages={1198--1203},
  year={2007},
  organization={IEEE}
}

@article{alexiadis2004next,
  title={The next generation simulation program},
  author={Alexiadis, Vassili and Colyar, James and Halkias, John and Hranac, Rob and McHale, Gene},
  journal={Institute of Transportation Engineers. ITE Journal},
  volume={74},
  number={8},
  pages={22},
  year={2004},
  publisher={Institute of Transportation Engineers}
}

@inproceedings{krajewski2018highd,
  title={The highd dataset: A drone dataset of naturalistic vehicle trajectories on german highways for validation of highly automated driving systems},
  author={Krajewski, Robert and Bock, Julian and Kloeker, Laurent and Eckstein, Lutz},
  booktitle={2018 21st international conference on intelligent transportation systems (ITSC)},
  pages={2118--2125},
  year={2018},
  organization={IEEE}
}

@inproceedings{bock2020ind,
  title={The ind dataset: A drone dataset of naturalistic road user trajectories at german intersections},
  author={Bock, Julian and Krajewski, Robert and Moers, Tobias and Runde, Steffen and Vater, Lennart and Eckstein, Lutz},
  booktitle={2020 IEEE Intelligent Vehicles Symposium (IV)},
  pages={1929--1934},
  year={2020},
  organization={IEEE}
}

@inproceedings{xu2022drone,
  title={SIND: A drone dataset at signalized intersection in China},
  author={Xu, Yanchao and Shao, Wenbo and Li, Jun and Yang, Kai and Wang, Weida and Huang, Hua and Lv, Chen and Wang, Hong},
  booktitle={2022 IEEE 25th International Conference on Intelligent Transportation Systems (ITSC)},
  pages={2471--2478},
  year={2022},
  organization={IEEE}
}

@inproceedings{caesar2020nuscenes,
  title={nuscenes: A multimodal dataset for autonomous driving},
  author={Caesar, Holger and Bankiti, Varun and Lang, Alex H and Vora, Sourabh and Liong, Venice Erin and Xu, Qiang and Krishnan, Anush and Pan, Yu and Baldan, Giancarlo and Beijbom, Oscar},
  booktitle={Proceedings of the IEEE/CVF conference on computer vision and pattern recognition},
  pages={11621--11631},
  year={2020}
}

@inproceedings{sun2020scalability,
  title={Scalability in perception for autonomous driving: Waymo open dataset},
  author={Sun, Pei and Kretzschmar, Henrik and Dotiwalla, Xerxes and Chouard, Aurelien and Patnaik, Vijaysai and Tsui, Paul and Guo, James and Zhou, Yin and Chai, Yuning and Caine, Benjamin and others},
  booktitle={Proceedings of the IEEE/CVF conference on computer vision and pattern recognition},
  pages={2446--2454},
  year={2020}
}

@inproceedings{chang2019argoverse,
  title={Argoverse: 3d tracking and forecasting with rich maps},
  author={Chang, Ming-Fang and Lambert, John and Sangkloy, Patsorn and Singh, Jagjeet and Bak, Slawomir and Hartnett, Andrew and Wang, De and Carr, Peter and Lucey, Simon and Ramanan, Deva and others},
  booktitle={Proceedings of the IEEE/CVF conference on computer vision and pattern recognition},
  pages={8748--8757},
  year={2019}
}

@article{wilson2023argoverse,
  title={Argoverse 2: Next generation datasets for self-driving perception and forecasting},
  author={Wilson, Benjamin and Qi, William and Agarwal, Tanmay and Lambert, John and Singh, Jagjeet and Khandelwal, Siddhesh and Pan, Bowen and Kumar, Ratnesh and Hartnett, Andrew and Pontes, Jhony Kaesemodel and others},
  journal={arXiv preprint arXiv:2301.00493},
  year={2023}
}

@inproceedings{houston2021one,
  title={One thousand and one hours: Self-driving motion prediction dataset},
  author={Houston, John and Zuidhof, Guido and Bergamini, Luca and Ye, Yawei and Chen, Long and Jain, Ashesh and Omari, Sammy and Iglovikov, Vladimir and Ondruska, Peter},
  booktitle={Conference on Robot Learning},
  pages={409--418},
  year={2021},
  organization={PMLR}
}

@inproceedings{xiao2021pandaset,
  title={Pandaset: Advanced sensor suite dataset for autonomous driving},
  author={Xiao, Pengchuan and Shao, Zhenlei and Hao, Steven and Zhang, Zishuo and Chai, Xiaolin and Jiao, Judy and Li, Zesong and Wu, Jian and Sun, Kai and Jiang, Kun and others},
  booktitle={2021 IEEE international intelligent transportation systems conference (ITSC)},
  pages={3095--3101},
  year={2021},
  organization={IEEE}
}

@inproceedings{yu2022dair,
  title={Dair-v2x: A large-scale dataset for vehicle-infrastructure cooperative 3d object detection},
  author={Yu, Haibao and Luo, Yizhen and Shu, Mao and Huo, Yiyi and Yang, Zebang and Shi, Yifeng and Guo, Zhenglong and Li, Hanyu and Hu, Xing and Yuan, Jirui and others},
  booktitle={Proceedings of the IEEE/CVF Conference on Computer Vision and Pattern Recognition},
  pages={21361--21370},
  year={2022}
}

@inproceedings{huang2018apolloscape,
  title={The apolloscape dataset for autonomous driving},
  author={Huang, Xinyu and Cheng, Xinjing and Geng, Qichuan and Cao, Binbin and Zhou, Dingfu and Wang, Peng and Lin, Yuanqing and Yang, Ruigang},
  booktitle={Proceedings of the IEEE conference on computer vision and pattern recognition workshops},
  pages={954--960},
  year={2018}
}

@inproceedings{geiger2012CVPR,
  author = {Andreas Geiger and Philip Lenz and Raquel Urtasun},
  title = {Are we ready for Autonomous Driving? The KITTI Vision Benchmark Suite},
  booktitle = {Conference on Computer Vision and Pattern Recognition (CVPR)},
  year = {2012}
}

@inproceedings{Qiu2019driving,
author = {Qiu, Yuning and Misu, Teruhisa and Busso, Carlos},
title = {Driving Anomaly Detection with Conditional Generative Adversarial Network using Physiological and CAN-Bus Data},
year = {2019},
isbn = {9781450368605},
publisher = {Association for Computing Machinery},
address = {New York, NY, USA},
url = {https://doi.org/10.1145/3340555.3353749},
doi = {10.1145/3340555.3353749},
booktitle = {2019 International Conference on Multimodal Interaction},
pages = {164–173},
numpages = {10},
location = {Suzhou, China},
series = {ICMI '19}
}

@inproceedings{Ramanishka_2018_CVPR,
author = {Ramanishka, Vasili and Chen, Yi-Ting and Misu, Teruhisa and Saenko, Kate},
title = {Toward Driving Scene Understanding: A Dataset for Learning Driver Behavior and Causal Reasoning},
booktitle = {Proceedings of the IEEE Conference on Computer Vision and Pattern Recognition (CVPR)},
month = {June},
year = {2018}
}

@misc{lyft2019,
title = {Lyft Level 5 AV Dataset 2019},
author = {Kesten, R. and Usman, M. and Houston, J. and Pandya, T. and Nadhamuni, K. and Ferreira, A. and Yuan, M. and Low, B. and Jain, A. and Ondruska, P. and Omari, S. and Shah, S. and Kulkarni, A. and Kazakova, A. and Tao, C. and Platinsky, L. and Jiang, W. and Shet, V.},
year = {2019}
}

@inproceedings{pham20203d,
  title={A* 3d dataset: Towards autonomous driving in challenging environments},
  author={Pham, Quang-Hieu and Sevestre, Pierre and Pahwa, Ramanpreet Singh and Zhan, Huijing and Pang, Chun Ho and Chen, Yuda and Mustafa, Armin and Chandrasekhar, Vijay and Lin, Jie},
  booktitle={2020 IEEE International conference on Robotics and Automation (ICRA)},
  pages={2267--2273},
  year={2020},
  organization={IEEE}
}

@article{geyer2020a2d2,
  title={A2d2: Audi autonomous driving dataset},
  author={Geyer, Jakob and Kassahun, Yohannes and Mahmudi, Mentar and Ricou, Xavier and Durgesh, Rupesh and Chung, Andrew S and Hauswald, Lorenz and Pham, Viet Hoang and M{\"u}hlegg, Maximilian and Dorn, Sebastian and others},
  journal={arXiv preprint arXiv:2004.06320},
  year={2020}
}

@article{zhang2002flexible,
  title={A flexible new technique for camera calibration},
  author={Zhang, Zhengyou},
  journal={IEEE Transactions on pattern analysis and machine intelligence},
  volume={22},
  number={11},
  pages={1330--1334},
  year={2002},
  publisher={IEEE}
}

@inproceedings{lang2019pointpillars,
  title={Pointpillars: Fast encoders for object detection from point clouds},
  author={Lang, Alex H and Vora, Sourabh and Caesar, Holger and Zhou, Lubing and Yang, Jiong and Beijbom, Oscar},
  booktitle={Proceedings of the IEEE/CVF conference on computer vision and pattern recognition},
  pages={12697--12705},
  year={2019}
}

@article{weng2020ab3dmot,
  title={Ab3dmot: A baseline for 3d multi-object tracking and new evaluation metrics},
  author={Weng, Xinshuo and Wang, Jianren and Held, David and Kitani, Kris},
  journal={arXiv preprint arXiv:2008.08063},
  year={2020}
}

@article{kalman1960new,
  title={A new approach to linear filtering and prediction problems},
  author={Kalman, Rudolph Emil},
  year={1960}
}

@inproceedings{wei2025integrating,
  title={Integrating multi-modal sensors: a review of fusion techniques for intelligent vehicles},
  author={Wei, Chuheng and Qin, Ziye and Zhang, Ziyan and Wu, Guoyuan and Barth, Matthew J},
  booktitle={2025 IEEE Intelligent Vehicles Symposium (IV)},
  pages={1817--1824},
  year={2025},
  organization={IEEE}
}

@article{wang2026drivedna,
  title={DriveDNA: A Large-Scale Multimodal Naturalistic Driving Dataset and Benchmark for Driving Style Identification},
  author={Wang, Yuhang and Li, Lingyao and Zhou, Hao},
  journal={arXiv preprint arXiv:2607.23822},
  year={2026}
}

@inproceedings{salzmann2020trajectron,
  title={Trajectron++: Dynamically-feasible trajectory forecasting with heterogeneous data},
  author={Salzmann, Tim and Ivanovic, Boris and Chakravarty, Punarjay and Pavone, Marco},
  booktitle={European conference on computer vision},
  pages={683--700},
  year={2020},
  organization={Springer}
}

@article{vaswani2017attention,
  title={Attention is all you need},
  author={Vaswani, Ashish and Shazeer, Noam and Parmar, Niki and Uszkoreit, Jakob and Jones, Llion and Gomez, Aidan N and Kaiser, {\L}ukasz and Polosukhin, Illia},
  journal={Advances in neural information processing systems},
  volume={30},
  year={2017}
}

@article{graves2012long,
  title={Long short-term memory},
  author={Graves, Alex},
  journal={Supervised sequence labelling with recurrent neural networks},
  pages={37--45},
  year={2012},
  publisher={Springer}
}

@inproceedings{schubert2008comparison,
  title={Comparison and evaluation of advanced motion models for vehicle tracking},
  author={Schubert, Robin and Richter, Eric and Wanielik, Gerd},
  booktitle={2008 11th international conference on information fusion},
  pages={1--6},
  year={2008},
  organization={IEEE}
}

@inproceedings{alahi2016social,
  title={Social lstm: Human trajectory prediction in crowded spaces},
  author={Alahi, Alexandre and Goel, Kratarth and Ramanathan, Vignesh and Robicquet, Alexandre and Fei-Fei, Li and Savarese, Silvio},
  booktitle={Proceedings of the IEEE conference on computer vision and pattern recognition},
  pages={961--971},
  year={2016}
}

@article{liu2024survey,
  title={A survey on autonomous driving datasets: Statistics, annotation quality, and a future outlook},
  author={Liu, Mingyu and Yurtsever, Ekim and Fossaert, Jonathan and Zhou, Xingcheng and Zimmer, Walter and Cui, Yuning and Zagar, Bare Luka and Knoll, Alois C},
  journal={IEEE Transactions on Intelligent Vehicles},
  volume={9},
  number={11},
  pages={7138--7164},
  year={2024},
  publisher={IEEE}
}

@INPROCEEDINGS{wei2026beyond,
  author={Wei, Chuheng and Cheng, Hao and Zhang, Ziyan and Wu, Guoyuan and Wang, Ziran and Barth, Matthew J.},
  booktitle={2026 IEEE Intelligent Vehicles Symposium (IV)}, 
  title={Beyond Perception: A Survey and Future Directions of VLM-based Autonomous Driving Datasets}, 
  year={2026},
  volume={},
  number={},
  pages={642-649},
  doi={10.1109/IV66570.2026.11624064}}

@article{wang2023multi,
  title={Multi-modal 3D object detection in autonomous driving: A survey and taxonomy},
  author={Wang, Li and Zhang, Xinyu and Song, Ziying and Bi, Jiangfeng and Zhang, Guoxin and Wei, Haiyue and Tang, Liyao and Yang, Lei and Li, Jun and Jia, Caiyan and others},
  journal={IEEE Transactions on Intelligent Vehicles},
  volume={8},
  number={7},
  pages={3781--3798},
  year={2023},
  publisher={IEEE}
}

@article{macenski2022robot,
  title={Robot operating system 2: Design, architecture, and uses in the wild},
  author={Macenski, Steven and Foote, Tully and Gerkey, Brian and Lalancette, Chris and Woodall, William},
  journal={Science robotics},
  volume={7},
  number={66},
  pages={eabm6074},
  year={2022},
  publisher={American Association for the Advancement of Science}
}
}

\end{document}